\documentclass[sigconf, anonymous=false, review=false]{acmart}
\usepackage{amsmath,flushend}
\usepackage{booktabs}
\usepackage{amsfonts}
\usepackage{amsbsy}
\usepackage{bbm}
\usepackage{algorithm}
\usepackage{algorithmicx,tikz}
\usepackage{algcompatible}
\usepackage{bm}
\usepackage{enumitem}
\usepackage{lipsum}
\usepackage{multirow}
\usepackage[normalem]{ulem}

\usepackage{url}

\usepackage{graphicx}
\usepackage{subcaption}

\newcommand{\name}{DISCO}

\newcommand{\hide}[1]{}

\def\BibTeX{{\rm B\kern-.05em{\sc i\kern-.025em b}\kern-.08emT\kern-.1667em\lower.7ex\hbox{E}\kern-.125emX}}
    
\copyrightyear{2022}
\acmYear{2022}
\setcopyright{acmcopyright}
\acmConference[CIKM '22]{Proceedings of the 31st ACM
International Conference on Information and Knowledge Management}{October
17--21, 2022}{Atlanta, GA, USA}
\acmBooktitle{Proceedings of the 31st ACM International Conference on Information
and Knowledge Management (CIKM '22), October 17--21, 2022, Atlanta, GA, USA}
\acmPrice{15.00}
\acmDOI{10.1145/3511808.3557202}
\acmISBN{978-1-4503-9236-5/22/10}

\begin{document}

\title{\name: Comprehensive and Explainable \\Disinformation Detection}

\author{Dongqi Fu}
\affiliation{\institution{University of Illinois at Urbana-Champaign}
  \state{Illinois}
  \country{USA}}
\email{dongqif2@illinois.edu}

\author{Yikun Ban}
\affiliation{\institution{University of Illinois at Urbana-Champaign}
  \state{Illinois}
  \country{USA}}
\email{yikunb2@illinois.edu}

\author{Hanghang Tong}
\affiliation{\institution{University of Illinois at Urbana-Champaign}
  \state{Illinois}
  \country{USA}}
\email{htong@illinois.edu}

\author{Ross Maciejewski}
\affiliation{\institution{Arizona State University}
\state{Arizona}
\country{USA}}
\email{rmacieje@asu.edu}

\author{Jingrui He}
\affiliation{\institution{University of Illinois at Urbana-Champaign}
  \state{Illinois}
  \country{USA}}
\email{jingrui@illinois.edu}

\begin{abstract}
Disinformation refers to false information deliberately spread to influence the general public, and the negative impact of disinformation on society can be observed in numerous issues, such as political agendas and manipulating financial markets. In this paper, we identify prevalent challenges and advances related to automated disinformation detection from multiple aspects and propose a comprehensive and explainable disinformation detection framework called \name. It leverages the heterogeneity of disinformation and addresses the opaqueness of prediction. Then we provide a demonstration of \name\ on a real-world fake news detection task with satisfactory detection accuracy and explanation. The demo video and source code of \name\ is now publicly available~\footnote{\url{https://github.com/DongqiFu/DISCO}}.
We expect that our demo could pave the way for addressing the limitations of identification, comprehension, and explainability as a whole.
\end{abstract}



\maketitle
\renewcommand{\shortauthors}{Dongqi Fu et al.} 

\section{Introduction}
Disinformation (e.g., fake news) is fabricated to mislead the general public. Historically, disinformation campaigns often took the form of government propaganda with edited newsreels, and the cost of creation and distribution required a funded and coordinated effort. However, with the recent rise of accessible and low-cost computational methods for text manipulation and the broad access to public information channels via the worldwide web, society now finds itself inundated with disinformation that is resulting in large-scale negative societal impacts.
Negative effects include but are not limited to, fake news deliberately misleads readers to accept false or biased information for further political agendas or manipulate financial markets; furthermore, fake news also downgrades the credibility of real news and hinders people's ability to distinguish factual information from disinformation~\cite{DBLP:journals/sigkdd/ShuSWTL17}.

Two main challenges hinder disinformation-related research.
Firstly, a characteristic of the spread and creation of disinformation is the co-existence of multiple types of heterogeneous features, which introduces ambiguous factors that could potentially camouflage disinformation from real information, i.e.,, even a single word can have different semantic meanings in different contexts~\cite{DBLP:conf/acl/SarmaLS18, DBLP:conf/emnlp/SarmaLS19}. For example, "Apple" in the food corpus means a kind of fruit, while it stands for the company in the high-tech corpus.
Secondly, interpretability is another key to understanding the logic of prediction systems to further build human-machine trust in results and improve model accuracy~\cite{DBLP:conf/kdd/ShuCW0L19,DBLP:conf/www/ZhouNMFH20, DBLP:journals/corr/abs-2110-14844}. 
While pioneering state-of-the-art algorithms have been proposed to detect disinformation, many of them are black-box in nature and lack interpretable mechanisms for explaining why information has been flagged as false.
Take fake news as an example, not every sentence in fake news is false. It is critical to specify guidelines for improving the existing model in an explainable way. For instance, to distinguish the importance of different sentences in determining the detection decision~\cite{DBLP:conf/kdd/ShuCW0L19}.
As such, there is a critical need for the new technology that can support disinformation detection both accurately and efficiently (at an early stage before widespread propagation), and in a way that is understandable by both domain experts and the general public.

The main contributions of the paper can be summarized as follows.
(1) We propose a comprehensive computational framework for disinformation, named \name. It models the heterogeneity of disinformation with graph machine learning techniques, such that the heterogeneity can be positively leveraged to solve challenges for detection and explanation.
To be specific, \textit{for the heterogeneity}, \name\ leverages large-scale pre-trained (from general corpus) language models in a transfer learning manner, which means involving a simple projection head could make \name\ achieve high detection accuracy in specific domains like politics news; \textit{for the explanation}, \name\ involves dynamics in a recently proposed graph neural network model~\cite{DBLP:conf/sigir/FuH21} such that the word and sentence importance towards the detection variance can be figured out by graph augmentations (i.e., masking nodes and edges in article graphs).
(2) We develop an online demo for visualizing the detection and the explanation result of \name. (3) We design the real-world experiment with 48,000+ fake and real news articles, and \name~ could outperform than baseline algorithms.

\textbf{Demonstration}.
The online demo of \name\ is programmed by Python and allows open interactions with users. The functions of the demo include (1) output the real and fake probabilities of a piece of suspect information; (2) output the misleading degree of each word in the input text and their rankings. A user-guide introduction video of the \name\ demo is also online now.

\textbf{Related Work}.
In current disinformation detection tools~\cite{DBLP:conf/cikm/ZhiSLZ017, DBLP:conf/www/PopatMSW18, DBLP:conf/www/MirandaNMVSGMM19, DBLP:conf/sigir/BotnevikSS20, DBLP:conf/cikm/0006RA21}, the query of suspicious information is usually a short phrase (e.g., "Microsoft is a Chinese company") and heavily relies on the retrieval of sufficient background articles to give a fair verdict to that query. To achieve the verification of long articles, the challenges include but not limited to the cost of searching time and searching scope, i.e., a long article may have many candidates from different aspects or searching an emerging disinformation may return a limited number of viable background articles. Our \name\ transfers the searching process into the question comprehension (i.e., via geometric feature extraction and neural detection), such that the open-end queries can be answered in the real time. Also, our model is built upon the large corpus, which has 3 million distinct words from numerous articles, can provide comprehensive knowledge for down-streaming specific domains with limited information. Among the above-mentioned demos~\cite{DBLP:conf/cikm/ZhiSLZ017, DBLP:conf/www/PopatMSW18, DBLP:conf/www/MirandaNMVSGMM19, DBLP:conf/sigir/BotnevikSS20, DBLP:conf/cikm/0006RA21}, our demo is most similar with the online module of~\cite{DBLP:conf/cikm/ZhiSLZ017}, which takes an open-end query and give the credibility analysis. Additionally, our \name\ also gives an interpretation in terms of each word’s contribution (positive or negative) towards the ground-truth prediction.
\vspace{-6mm}

\section{System Architecture}
\begin{figure}[t]
\includegraphics[width=0.45\textwidth]{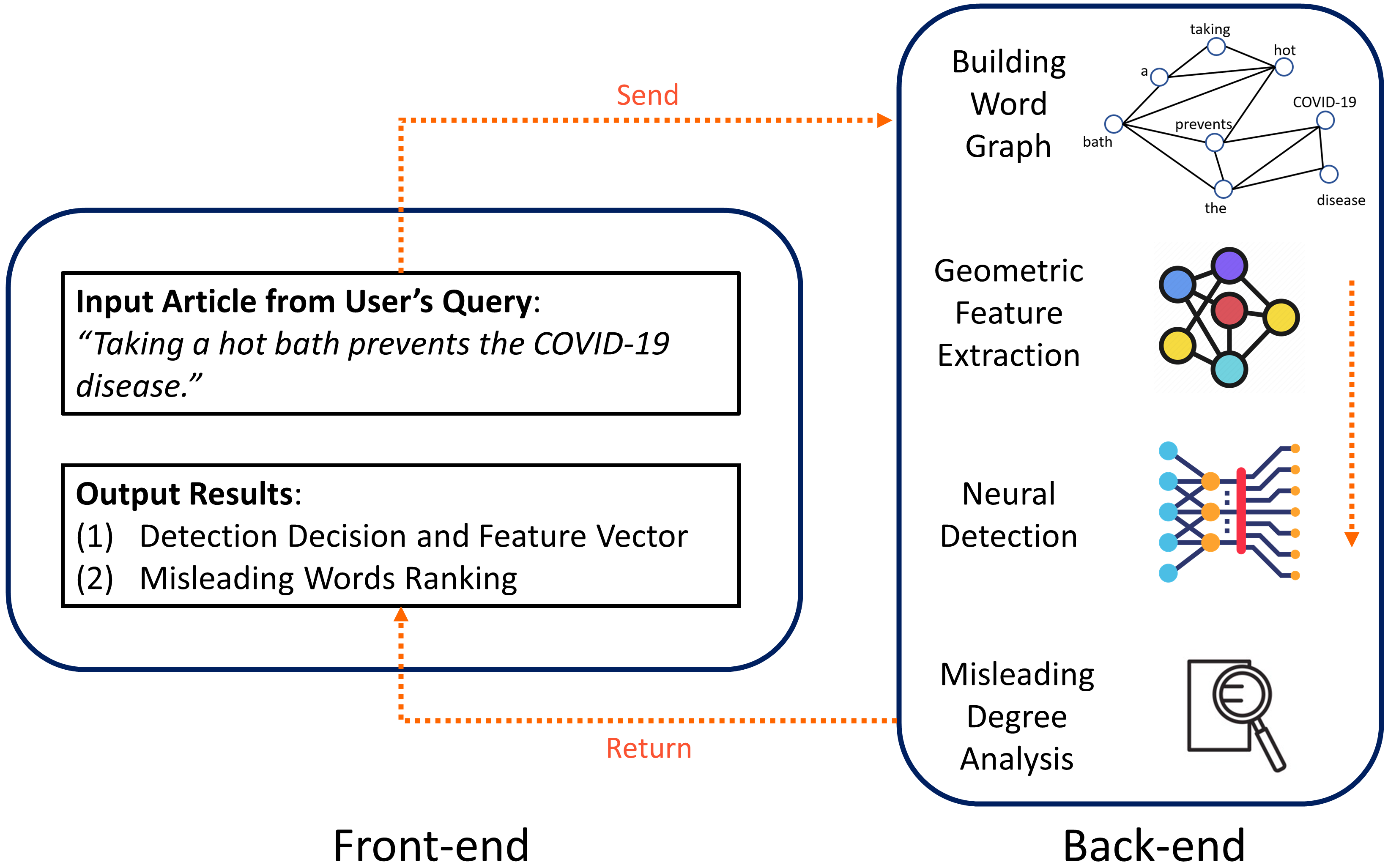}
\centering
\vspace{-3mm}
\caption{System Architecture of \name.}
\vspace{1mm}
\label{Fig:software_architecture}
\end{figure}

The online demo of \name\ has two main parts, front-end, and back-end, as shown in Figure~\ref{Fig:software_architecture}. The front-end (1) accepts and passes the user open query (i.e., a suspect piece of information) to the back-end, and (2) receives and shows the detection decision and misleading words rankings from the back-end. The back-end is supported by the graph machine learning techniques, which is responsible for identifying the input information and making the corresponding explanation by (1) building the word graph for the input article (2) extracting the geometric feature of that graph; (3) predicting whether the input article is real or fake; and (4) ranking each word in that text based on the misleading degree. Then the back-end returns all results to the front-end for users. The theoretical details of how these four sequential functions inside the back-end get realized are discussed in the next section.

\begin{figure*}[t]
\includegraphics[width=0.8\textwidth]{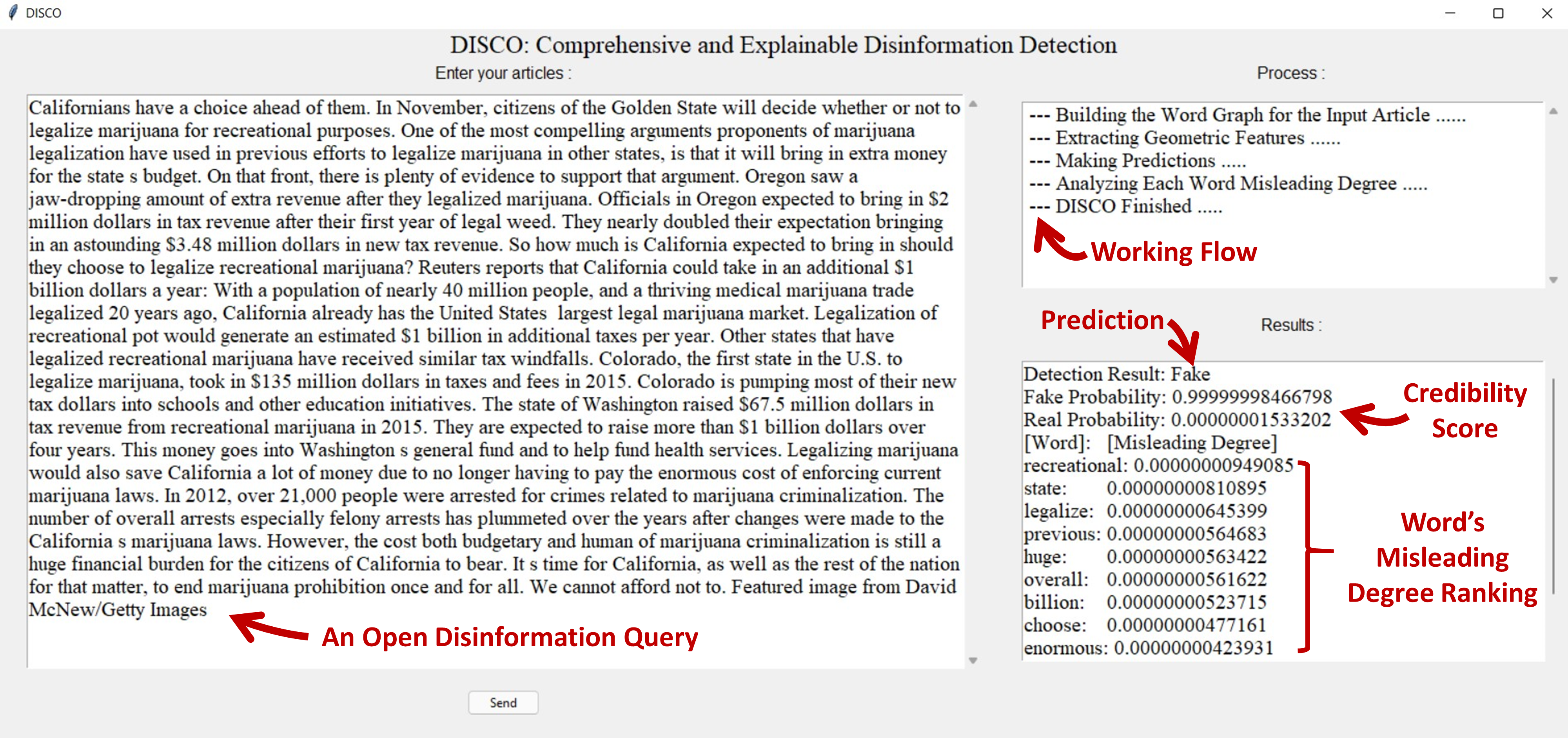}
\centering
\vspace{-3mm}
\caption{User Interface of DISCO. The left-hand side is the text area to receive open queries from users, the upper right-hand side is the real-time process monitor of \name, and the lower right-hand side is the prediction and explanation result.}
\label{Fig:user_interface}
\vspace{-4.5mm}
\end{figure*}

\section{Theoretical Details}
The back-end side of our \name\ demo is based on graph machine learning techniques, such that an article can be represented as an embedding vector, and the disinformation detection problem is converted into a graph classification problem.
First of all, an input article is modeled by a word graph (Subsection 3.1). Second, the entire graph is represented by an embedding vector (Subsection 3.2). Third, the graph-level embedding vector goes into a neural network to get the predictions in terms of the fake news probability and real news probability (Subsection 3.3). Fourth, each word in the article (i.e., each node in the built word graph) is masked under the same graph topological constraint to see each word's contribution towards the input article prediction (Subsection 3.4).

\vspace{-2mm}

\subsection{Building Word Graphs}
To build a word graph $G$ (i.e., upper right corner in Figure~\ref{Fig:software_architecture}) for an input news article, we follow~\cite{DBLP:conf/emnlp/MihalceaT04} such that each word in the article stands for a unique node and an edge is established if two words co-occur in a window of $k$ text units. In our demonstration, we set $k=3$. Take "I eat an apple" as an example, and then the edges could be \{I-eat, I-an, eat-an, eat-apple, an-apple\} with stop words kept and edges undirected. Beyond~\cite{DBLP:conf/emnlp/MihalceaT04}, we assign each node $i$ in graph $G$ with its own node feature vector $\bm{x}_{i} \in \mathbb{R}^{1 \times d}$, and $\bm{x}_{i}$ that can be retrieved from the word embedding vector of the large-scale pre-trained NLP model, like Bert~\cite{DBLP:conf/naacl/DevlinCLT19} or Word2Vec~\cite{DBLP:conf/nips/MikolovSCCD13}. In our demonstration, we adopt a large-scale pre-trained Word2Vec from Google~\footnote{\url{https://code.google.com/archive/p/word2vec/}} that is pre-trained on a 3 million distinct words corpus, and each word embedding vector is 300-dimensional (i.e., $d=300$).

\vspace{-2mm}

\subsection{Geometric Feature Extraction}
Suppose there are $n$ different words in an article, then we can construct a word graph $G$ with $n$ nodes as mentioned above. Given the input node feature matrix $\bm{X} \in \mathbb{R}^{n \times d}$ (i.e., $X(i,:) =\bm{x}_{i}$), the node hidden representation vector $\bm{h}_{i} \in \mathbb{R}^{1 \times d}$, $i \in \{1, \ldots, n\}$, can be obtained by
\vspace{-1mm}
\begin{equation}
    \bm{h}_{i} = \bm{p}_{i}^{\top} \bm{X} 
\label{eq:content representation}
\vspace{-0.5mm}
\end{equation}
where $\bm{p}_{i} \in \mathbb{R}^{n \times 1}$ is the personalized PageRank vector with node $i$ as the seed node and can be expressed as follows
\vspace{-1mm}
\begin{equation}
    \bm{p}_{i} = \alpha \bm{A}\bm{D}^{-1} \bm{p}_{i} + (1-\alpha)\bm{r}
\label{eq:stationary distribution}
\vspace{-1mm}
\end{equation}
where $\bm{A}$ and $\bm{D}$ are adjacency and degree matrices of word graph $G$, $\alpha \in [0,1]$ is the teleportation probability (e.g., $\alpha =0.85$ in our demo), and $\bm{r} \in \mathbb{R}^{n \times 1}$ is the personalized vector with $r(i) = 1$ and other entries are 0s.

With this geometric feature extraction, the heterogeneous semantic meanings of words are jointly modeled.
To be more specific, $\bm{p}_{i}$ encodes the stationary distribution of random walks starting from node $i$, it can be interpreted as the relevant weights of other nodes (i.e., words) to the seed node (i.e., selected word) in this graph $G$ (i.e., input news article). Our input node feature $\bm{x}_{i}$ is general because it is distilled by pre-trained models from a large-scale corpus like Wikipedia. Thus, according to Eq.~\ref{eq:content representation}, $\bm{x}_{i}$ is in-depth specialized by $\bm{h}_{i}$, which means the meaning of a selected word is specialized by its neighbor words in the scope of this input article.

Then, to obtain the graph-level (i.e., document-level) representation $\bm{u} \in \mathbb{R}^{1 \times d}$ for the word graph $G$ (i.e., input news article), we read out all word-level hidden representations $\bm{h}_{i}$ as follows
\vspace{-1mm}
\begin{equation}
    \bm{u} = readout(\bm{h}_{i} ~|~ i \in \{1, \ldots, n\})
\label{eq:readout}
\vspace{-1mm}
\end{equation}
where the $readout$ function is permutation-invariant and could be the graph pooling layer, such as sum or average pooling~\cite{DBLP:conf/aaai/ZhangCNC18, DBLP:conf/nips/YingY0RHL18}.

This geometric feature extraction is not only effective because it replaces the traditional message-passing scheme of stacking GNN layers with the stationary distribution based aggregation~\cite{DBLP:conf/iclr/KlicperaBG19, DBLP:conf/kdd/BojchevskiKPKBR20}; but is also efficient, which means the new stationary distribution can be fast tracked when the graph topology changes~\cite{DBLP:conf/kdd/FuZH20, DBLP:conf/cikm/FuXLTH20, DBLP:conf/sigir/FuH21, DBLP:journals/corr/abs-2107-02168, DBLP:conf/kdd/FuFMTH22, DBLP:conf/kdd/ZhouZ0H20, DBLP:journals/fdata/ZhouZXH19, DBLP:conf/www/JingTZ21, DBLP:conf/www/JingPT21, DBLP:conf/aaai/YanLBJT21, DBLP:conf/www/YanZT21, DBLP:journals/corr/abs-2110-13798}, e.g., in this demo, we can mask several nodes in the input word graph without retraining. These two merits pave the way for our explanation function of \name. 

\vspace{-2mm}

\subsection{Neural Detection}
The loss function of the proposed \name\ model is deployed on the representation vector $\bm{u}_{j}$ against the label information $\bm{y}_{j}$ (e.g., fake or real) of the $j$-th news article. To realize this, we need to call a neural network $\mathbb{N}$ to transform each $\bm{u}_{j}$ into $\bm{z}_{j}$. For instance, the cross-entropy between these two is expressed as follows
\vspace{-1mm}
\begin{equation}
    \mathcal{L} = - \sum_{j} \bm{y}_{j} ~\text{ln}~ \bm{z}_{j}
\label{eq:distill}
\vspace{-2mm}
\end{equation}
where $\bm{z}_{j} \in \mathbb{R}^{1 \times q}$ is the final output of neural network $\mathbb{N}$ w.r.t $\bm{u}_{j}$, and $\bm{y}_{j} \in \mathbb{R}^{1 \times q}$ denotes the ground truth label of the entity $j$.

Benefiting from our geometric feature extraction scheme, the deployment of neural network $\mathbb{N}$ can be model-agnostic~\cite{DBLP:conf/iclr/KlicperaBG19, DBLP:conf/kdd/BojchevskiKPKBR20, DBLP:conf/sigir/FuH21}, which means the feature extraction is independent of the neural detection, and we can apply various kinds of neural networks according to different settings. For example, in our online demo, we instance $\mathbb{N}$ with a simple 32*2 multi-layer perceptron (MLP) for achieving effective performance, as shown in Figure~\ref{Fig:user_interface}. Additionally, we also provide the contextual multi-armed bandits in the exploitation-exploration dilemma~\cite{DBLP:conf/kdd/BanHC21, ban2021convolutional}.

\vspace{-4mm}

\subsection{Explanation of Misleading Words}
Given an article is detected as fake or real, each word has different contributions to this decision. For example, some words help disinformation camouflage and hider the detection to detect it. After removing such words in that article, the decision can be more deterministic, i.e., the corresponding prediction probability increases. In this paper, we call these words "misleading words". Next, we explain how our \name\ could explain each word's misleading degree.

We can choose to mask any nodes in the word graph $G_{j}$ to see their contributions to the final representation $\bm{z}_{j}$, to further see to what extent the prediction is changed. Therefore, we can know what factors make DISCO dictate such a prediction, which is especially helpful for misclassified cases. Technically, masking nodes and edges without re-training the model from scratch relies on our proposed Eq.~\ref{eq:content representation} and Eq.~\ref{eq:readout}, where $\bm{p}_{i}$ is the stationary distribution of seed node $i$ on graph $G$. When we need to mask a certain node in graph $G$ and change it into $G'$, the new stationary distribution $\bm{p}_{i}'$ can be fast and accurately tracked only with the topology changes but without the neural network parameters fine-tuning~\cite{DBLP:conf/sigir/FuH21}.

In Eq.~\ref{eq:stationary distribution}, we use $\bm{M} = \bm{A}\bm{D}^{-1} \in \mathbb{R}^{n \times n}$ to denote the column-stochastic transition matrix. When a certain node is masked (i.e., its adjacent edges are deleted), the graph topology will change from $\bm{M}$ to $\bm{M}'$, and the new stationary distribution $\bm{p}_{i}'$ of each node $i$ needs to be tracked. To obtain each $\bm{p}_{i}'$, the core idea is to push out the previous probability distribution score from the changed part to the residual part of the graph $G$, and then add the pushed out distribution $\bm{p}_{i}^{pushout}$ back to the previous distribution $\bm{p}_{i}$ to finally obtain the new distribution $\bm{p}_{i}'$. The tracking process can be described as follows
\vspace{-2mm}
\begin{equation}
\label{eq:track_1}
    \bm{p}_{i}^{pushout} = \alpha (\bm{M}' - \bm{M}) \bm{p}_{i}
\end{equation}
\vspace{-4mm}
and
\begin{equation}
\label{eq:track_2}
    \bm{p}_{i}' = \bm{p}_{i} + \sum_{k=0}^{\infty}(\alpha\bm{M}')^{k}~\bm{p}_{i}^{pushout}
\vspace{-2mm}
\end{equation}
where $\bm{p}_{i}^{pushout}$ denotes the distribution score that needs to be pushed out on the residual graph due to the updated edges, and $\bm{p}_{i}'$ denotes the tracked new distribution. The above pushout process can be proved to converge to the exact stationary distribution of the new graph through sufficient cumulative power iterations~\cite{DBLP:conf/icde/YoonJK18,DBLP:conf/www/YoonJK18}.

After we get each new $\bm{p}_{i}'$, according to Eq.~\ref{eq:content representation} and Eq.~\ref{eq:readout}, we can get the new graph-level hidden representation $\bm{u}'$ and final representation $\bm{z}'$ without fine-tuning neural network $\mathbb{N}$. Then the difference between the correct prediction probability of $\bm{z}'$ and $\bm{z}$ composes the misleading degree of the masked word. For example, as shown in Figure~\ref{Fig:user_interface}, when we mask the word "recreational" in the input news article, the new probability of predicting this new article (i.e., $\bm{z}'$) as fake news is 99.999999415\%. Compared with 99.999998466\% of the original article (i.e., $\bm{z}$), the confidence of correct prediction increases 0.000000949\% by masking the word "recreational". Therefore, the word "recreational" hinders DISCO make the correct prediction, and its misleading degree is 0.00000000949. In our demo, we rank each word based on its misleading degree and show the rank in decreasing order. Also, the misleading degree can be negative, which means that the masked word in the original article helps DISCO make correct predictions.

\vspace{-2mm}

\section{Real-World Evaluation}

\vspace{-1mm}

\subsection{Experiment Preparation}

\textbf{Data Set}.
We choose the fake real news data set from~\cite{DBLP:conf/isddc/AhmedTS17, DBLP:journals/sap/AhmedTS18}, and each news article focusing on politics around the world has a title, text, subject, date, and a label indicating it is fake news or not. After preprocessing, the statistics of valid records of fake news and real news are stated as below. The fake news articles consist of 23,481 items, ranging from Mar 31, 2015 to Feb 19, 2018. As for real news articles, 21,417 items are involved from Jan 13, 2016 to Dec 31, 2017.
\textbf{Environment Setup}.
First, we initialize Google pre-trained Word2Vec for the node input feature $\bm{X}$ in Eq.~\ref{eq:content representation}. Second, we use the sum-pooling function as the $readout$ function in Eq.~\ref{eq:readout}. Third, we use a multi-layer perceptron with the ReLU activation function to distill $\bm{u}$ in Eq.~\ref{eq:distill}, where the optimizer is based on Adam and the initial learning rate is set to be 0.0001.

\vspace{-2mm}

\subsection{Performance of \name}
\begin{figure}[h]
\includegraphics[width=0.35\textwidth]{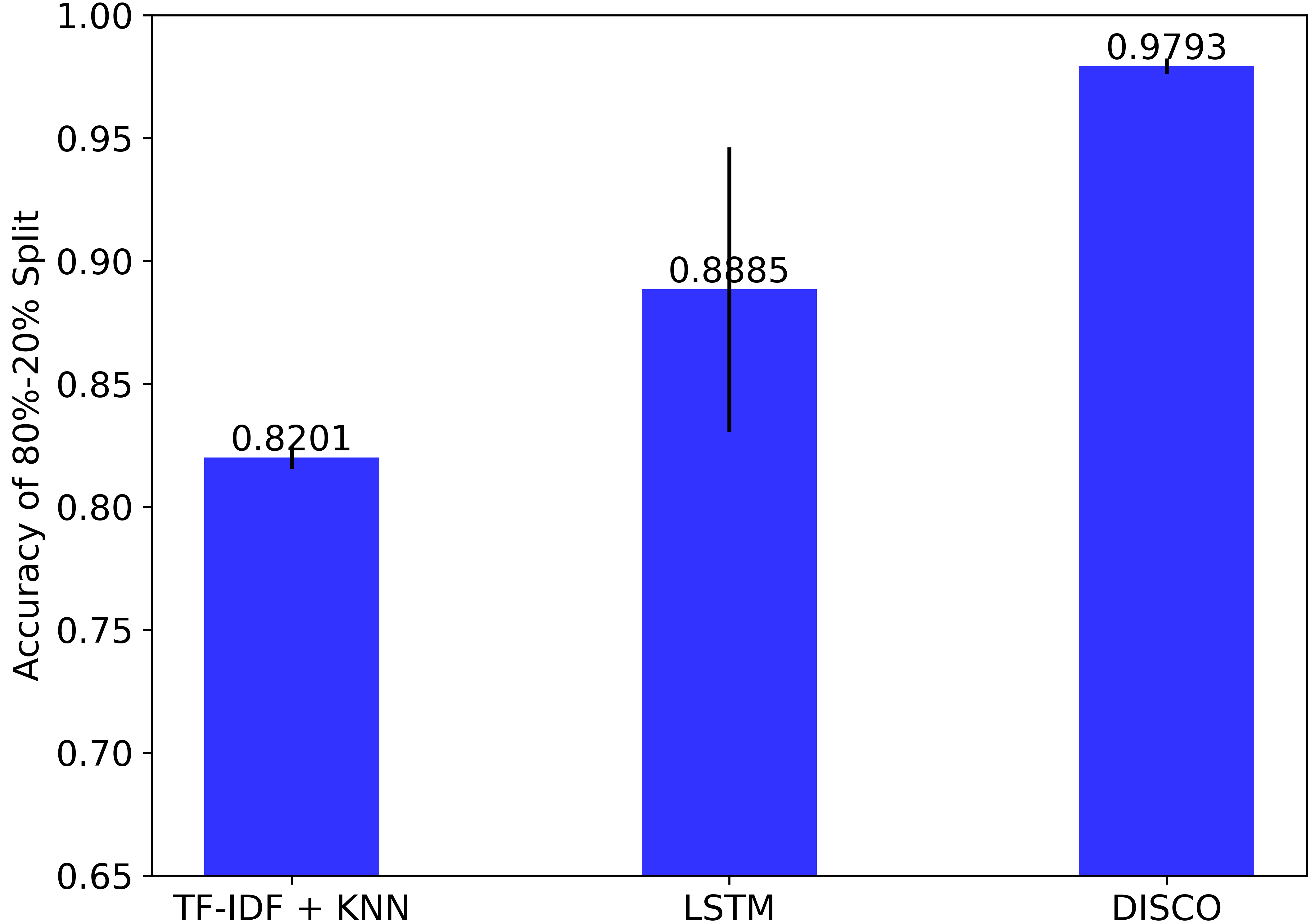}
\centering
\vspace{-3mm}
\caption{Performance on the 80\%-20\% Split Setting.}
\label{Fig:performance}
\vspace{-4mm}
\end{figure}

\textbf{Accuracy}. We show the performance of different baselines with our \name\ and report their fake news detection accuracy in Figure~\ref{Fig:performance}. For each bar in Figure~\ref{Fig:performance}, we report the performance w.r.t the average and standard deviation under the $80\%$--$20\%$ training-testing split 10 times randomly. From Figure~\ref{Fig:performance}, we can see that \name\ achieves the best performance. A possible interpretation is that the graph modeling of \name\ is better to model the relationship between words in articles than sequential modeling (i.e., BOW + LSTM) or non structural modeling (i.e., TF-IDF + $k$NN, $k=3$).

\begin{figure}[h]
\includegraphics[width=0.35\textwidth]{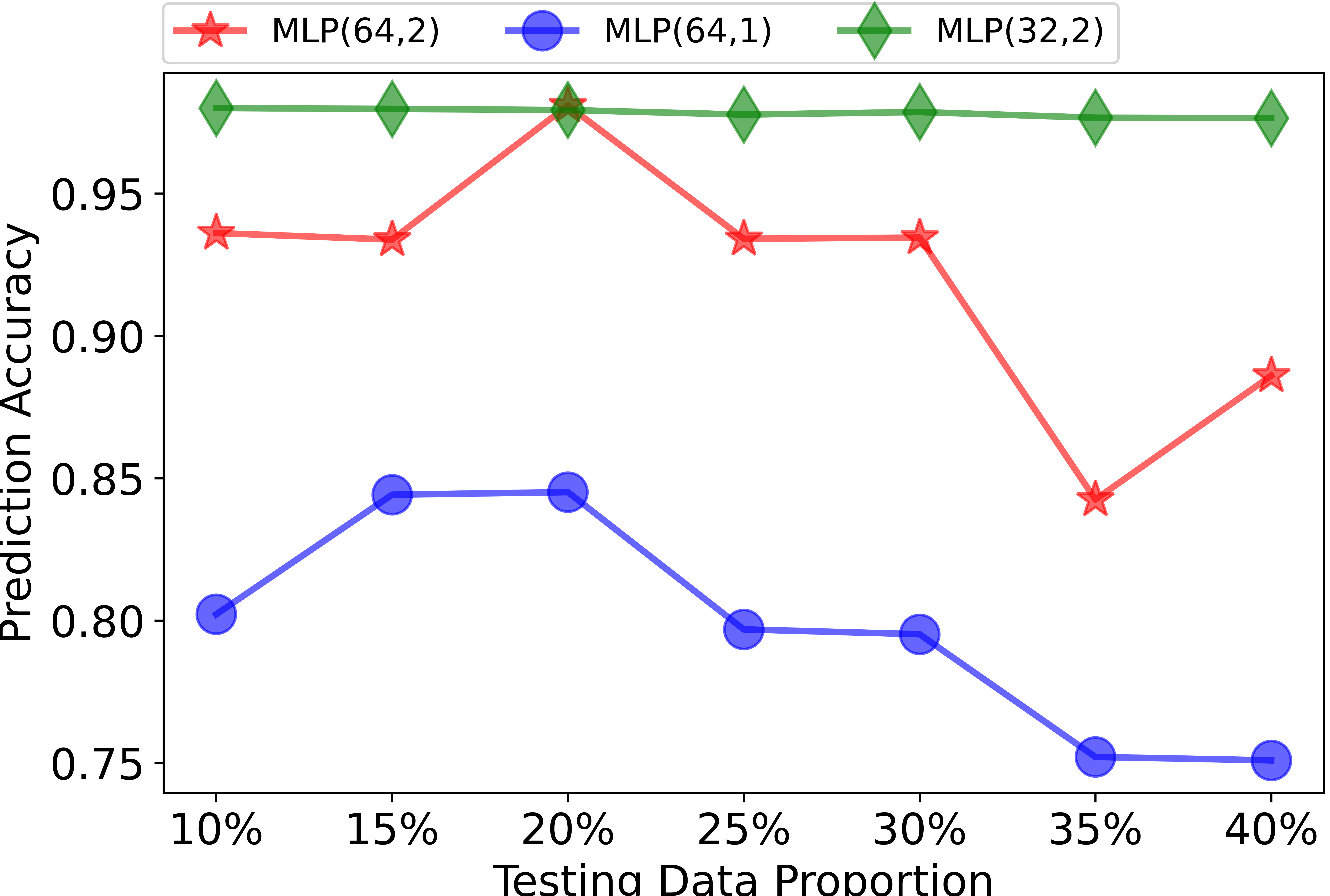}
\centering
\vspace{-3mm}
\caption{Performance of \name\ on Different Settings.}
\label{Fig:albation}
\vspace{-6mm}
\end{figure}

\textbf{Robustness}.
From the ablation study as shown in Figure~\ref{Fig:albation}, we can observe that (1) even the sample neural detection module can achieve competitive performance under the \name\ modeling, for MLP (hidden dim = 32, \# layers = 2) achieving the most accurate and stable predictions; (2) in many cases, too few or too many training samples could not contribute the most to the neural detection module under our \name\ modeling, for MLP (hidden dim = 64, \# layers = 1) and MLP (hidden dim = 64, \# layers = 2) reaching the peak when $20\%$ of the whole data set are extracted as testing samples.

\vspace{-2mm}

\textbf{Sensitivity}. From Figure~\ref{Fig:albation}, we know that cooperating with MLP (hidden dim = 32, \# layers = 2) our \name\ is robust for the low variance with varying testing sample sizes. Here, we want to investigate the prediction sensitivity of \name. Again, based on the 75\%--25\% training-testing split, we shuffle the entire data set 10 times randomly and report the precision, recall, and F1-score. The precision of \name\ is $0.9748 \pm 0.0086$, which means that in all the positive predictions, 97.48\% of them are correct predictions. The recall of \name\ is $0.9754 \pm 0.0065$, which means that among all ground-truth positive items, 97.54\% of them are identified by \name. The average and standard deviation of the F1-score is $0.9750 \pm 0.0010$, and the low standard deviation suggests that each time we achieve high precision, we also achieve a high recall simultaneously.

\vspace{-3mm}

\section{Conclusion}
\vspace{-1mm}
In this paper, we identify disinformation detection challenges and make an attempt for proposing \name\ and demonstrate it.
We wish that the next generation of disinformation detection systems could be able to simultaneously detect and explain during the whole life cycle of disinformation dissemination. 
\vspace{-2mm}


\section*{Acknowledgement}
\vspace{-1mm}
This work is supported by the National Science Foundation (Award Number IIS-1947203, IIS-2117902, and IIS-2137468), and the U.S. Department of Homeland Security (Award Number 17STQAC00001-05-00). The views and conclusions are those of the authors and should not be interpreted as representing the official policies of the funding agencies or the government.

\bibliographystyle{ACM-Reference-Format}
\bibliography{reference.bib}
\end{document}